\documentclass{article}
\usepackage{siunitx}
\usepackage{arxiv}

\usepackage[utf8]{inputenc} 
\usepackage[T1]{fontenc}    
\usepackage{hyperref}       
\usepackage{url}            
\usepackage{booktabs}       
\usepackage{amsfonts}       
\usepackage{nicefrac}       
\usepackage{microtype}      
\usepackage{lipsum}		
\usepackage{graphicx}
\usepackage{natbib}
\usepackage{doi}

\usepackage{graphicx}

\title{ Personality-Driven Gaze Animation with Conditional  Generative Adversarial Networks}

\author{ 
	\href{https://orcid.org/0000-0002-4915-6642} Funda Durupinar\\
	Department of Computer Science\\
	University of Massachusetts Boston\\
	Boston, Massachusetts \\
	\texttt{funda.durupinarbabur@umb.edu} \\
	}
	
\begin{document}
\maketitle

\begin{abstract}
We present a generative adversarial learning approach to synthesize gaze behavior of a given personality. We train the model using an existing data set that comprises eye-tracking data and personality traits of 42 participants performing an everyday task. Given the values of Big-Five personality traits (openness, conscientiousness, extroversion, agreeableness, and neuroticism), our model generates time series data consisting of gaze target, blinking times, and pupil dimensions.  We use the generated data to synthesize the gaze motion of virtual agents on a game engine.
\end{abstract}








\section{Introduction}

Expressive eye movements are essential components of believable virtual character animation.  They effectively communicate  attention in addition to giving information about the emotional and mental states of the individual.
There are a number of factors that control and explain the various manners of gaze behavior, such as turn taking, information processing and scene context. Personality is one such factor. Studies suggest that there are correlations between different aspects of personality and gaze parameters such as gaze shifts and blink rates~\cite{Libby73, Rauthmann2012, Hoppe18, Berkovsky19}. 

In this work, we propose a data-driven, generative approach to synthesize gaze behaviors for different personalities. We use data acquired from individuals in an everyday setting as opposed to data from actors playing a given role based on known personality-gaze correlations~\cite{Ruhland2015}. This helps capture the small details of gaze cues not yet conceptualized, but reflecting certain personality traits. 

We employ the widely accepted Big-Five model of personality, which describes personality in five orthogonal dimensions of openness, conscientiousness, extroversion, agreeableness, and neuroticism~\cite{Goldberg90}. We train a generative adversarial network conditioned on personality classes for each dimension. Our model learns from an existing personality-annotated dataset by~\cite{Hoppe18}. The dataset includes participants' Big-Five values and time-series data for gaze coordinates, blinking times, and pupil dimensions.  We then use the generated data to animate the eye movements of a virtual model. 

Although deep learning has been used to generate gaze movement~\cite{Klein2019}, the applications are limited to eye and body pose coordination for target following. To our knowledge, our method is the first to apply deep learning to generate eye movement data based on personality expression.

\section{Related Work}

Gaze movement research combines knowledge from various disciplines including psychology, neuroscience, social sciences, machine learning and computer graphics.  In computer graphics, gaze behavior is animated both by procedural~\cite{Lance08, Peters10, Pejsa13} and data-driven approaches~\cite{Pejsa2016, Klein2019}.


With the recent advances in eye tracking and machine learning, correlations between features related to blinks and eyeball movements such as fixations and saccades with the Big Five personality traits have been established ~\cite{Rauthmann2012, Hoppe18, Berkovsky19}.  Gaze animation methods that reflect personality traits apply such correlations to create the desired effects ~\cite{Ruhland2015, Fukuyama2002}.

Generative Adversarial Networks (GANs) have been highly successful at synthesizing realistic data. Since their introduction by Goodfellow et al.~\cite{Goodfellow14}, many variations to capture specific conditions have been proposed~\cite{Radford2016, Arjovsky2017}. In addition to the original image synthesis domain, they have been applied to generating natural language~\cite{Yu2017}, time-series data such as health records~\cite{Hyland17}, music~\cite{Mogren16}, human motion~\cite{Barsoum2018, Ferstl2019}  and so on.  GANs are also popular in  computer graphics, especially for facial animation, locomotion and gesture synthesis~\cite{ Vougioukas2018,Sadoughi2019, Wang2019,  Ferstl2019}.


\section{Method}

\subsection{Data}
For training, we use the dataset provided by Hoppe et al.~\cite{Hoppe18}. The dataset consists of binocular eye movement data of 42 participants, each with an average of 12.51 minutes of tracking information and personality scores for five factors binned into three groups of low, medium and high. The data was acquired by head-mounted eye trackers while participants walked around the campus and purchased an item of their choice from a campus shop. The data was sampled at 60 Hz.  The availability of personality information and the everyday nature of the performed tasks as opposed to laboratory-confined tests make this dataset a good fit for our goals.

\subsection{Gaze Parameter Synthesis}
To synthesize gaze parameters, we build a Generative Adversarial Network (GAN) conditioned on personality values. The GAN is composed of two competing networks:  a discriminator ($D$) and a generator ($G$).  G learns a distribution $p_g$ over data $x$ while D is trained to discriminate between the real data and synthetic data $G(z)$, where $z$ is input noise drawn from a random normal distribution.   The two networks are trained simultaneously.  The problem can be identified as a minimax game with a value function $V(D, G)$ (~\cite{Goodfellow14}):

\begin{equation}
    min_G max_D V(D, G) = \mathbb{E}_{x\sim p_r}[log(D(x))] + \mathbb{E}_{z\sim p_g}[1-log(D(G(z))]
\end{equation}

Conditional GANs~\cite{Arnelid19, Mirza2014} condition the model on given classes, allowing direct data generation given class labels. In our model, we integer encode the personality values into class labels. 
The GAN is extended to handle conditional labels $y$ with distribution $p_l$ as:

\begin{equation}
    min_G max_D V(D, G)  = \mathbb{E}_{x\sim p_r, y \sim p_l}[log(D(x, y))]  + \mathbb{E}_{z\sim p_g, y \sim p_l}[1-log(D(G(z, y), y)]
\end{equation}

There are different models to make predictions on time series data. For instance, recurrent neural networks (RNN) perform well to model sequence data, especially with Long short-term memory (LSTM) units which are effective in learning longer dependencies.  1D convolutional neural network (CNN) models are also used for time-series data prediction as they allow time-invariant feature extraction.  They can also be easily extended to multivariate time-series data as in our case. In our experiments, we found deep CNNs (similar to the DCGAN architecture by ~\cite{Radford2016}) to perform better in terms of stability than RNNs with multiple LSTM units. 







\begin{figure}
  \includegraphics[width=0.9\textwidth]{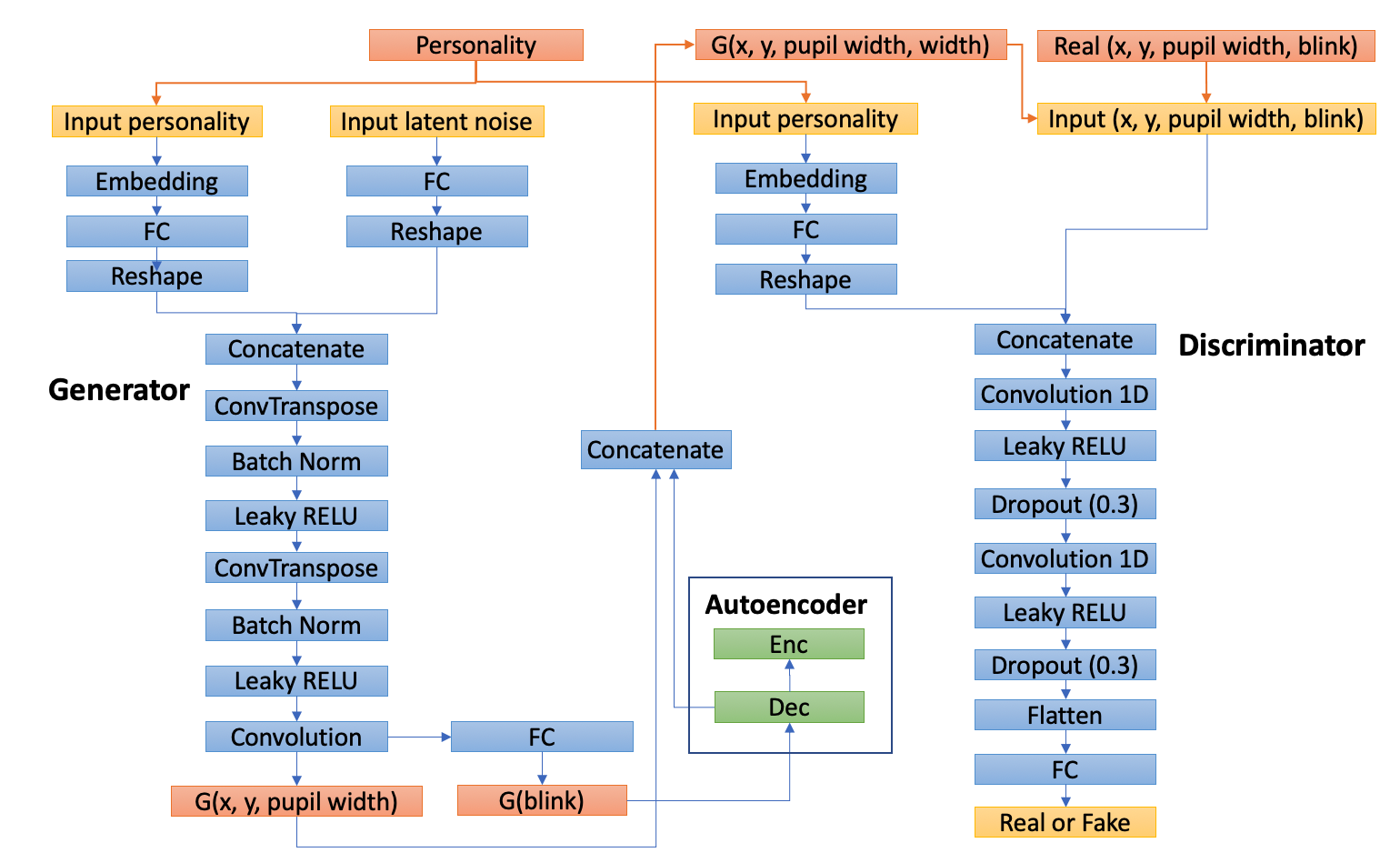}
  \caption{Conditional GAN architecture.}
  \label{fig:gan}
\end{figure}

The data features include both continuous and discrete parameters.  Gaze coordinates and pupil dimensions are continuous, and they can be directly fed into the GAN. However, blinking information is categorical. At each time step, it is zero if no blink was detected, or one if blink was detected.  Categorical data includes an additional step between the generator and the discriminator to provide continuous gradients. For this, we pre-train an autoencoder to encode and decode the binary data, and send the output of the generator to the decoder first, then send the output of the decoder to the discriminator. The model architecture is shown in Figure~\ref{fig:gan}.

\subsection{Model Parameters}
Discriminator input is a 4-dimensional vector comprising the x and y coordinates of gaze positions, the average pupil diameter of the left and the binary blinking data.   We organize the data by sliding windows of size 300 corresponding to 5 seconds of data at 60 Hz. For each of these 300-frame windows, we perform a strict quality test and discard the windows that include at least one row with x or y coordinates beyond the $[0, 1]$ range or pupil dimensions equal to zero. This leaves only the valid data points. Before feeding the continuous data into the GAN, we normalize it into the $[-1, 1]$ range.

Personality comprises the conditional class label that specifies each participant. It  is introduced to the network as a 50-dimensional embedding vector that encodes 243 possible values ($3^5$ for each bin and personality dimension).  Since the dataset is limited, only 24 out 243 possible classes are represented in the training set. This conditions the discriminator on the \emph{seen} classes, but allows the generator to predict sequences for \emph{unseen} classes. 

We also train the model for each personality dimension separately, where the personality dimension has three labels representing the low, medium and high values per personality.

We train mini-batches of size 64, using Adam optimizer  with a learning rate of 0.0001 both for D and G. The implementation is done using Keras functional API.


\subsection{Evaluation}

To evaluate the GAN model quantitatively, we train a deep 1-D CNN classifier on real data, synthesize a large number of data points and predict the probability of them belonging to each personality bin (class). Inception score is a metric to summarize these predictions~\cite{salimans16}.  Table~\ref{tab:inception} shows the scores for the test data for real and synthetic values.  When we train the classifier on all the five dimensions, because the representation in the training data is limited, the inception score is low.  We also compute the inception scores when each dimension is introduced as a condition separately. Considering that there are three classes per personality dimension, the closer the inception score to 3, the better the results.  We see that extroversion performs the best and neuroticism the worst.



\begin{table} [h]
    \centering
    \caption{Inception scores per personality for synthetic and real data}
    \begin{tabular}{c|c|c|c|c|c|c}
         Data &O & C &E &A  &N  & All dims \\
         \hline
         Synthetic &2.38& 2.25 & 2.56 & 2.41 & 2.56 & 6.23\\
         Real test &2.87 & 2.63 & 2.88 & 2.78 & 2.89 & 15.62
    \end{tabular}\label{tab:inception}
\end{table}

\begin{figure*}
  \includegraphics[width=\textwidth]{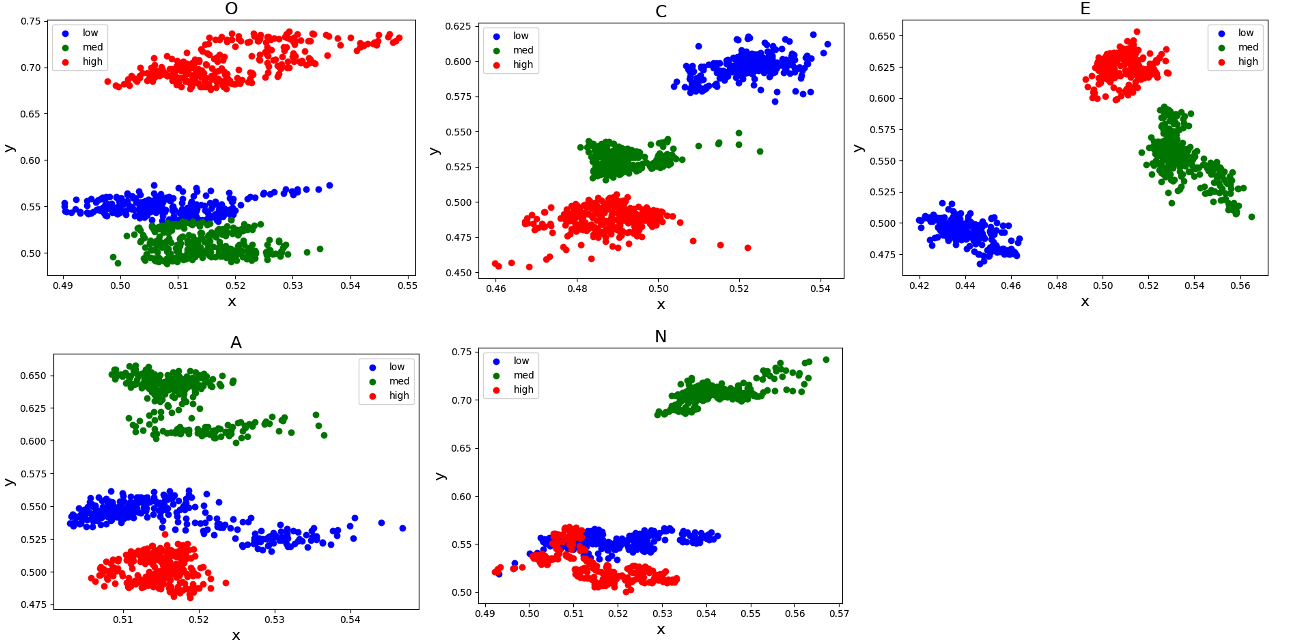}
  \caption{Synthesized average x and y coordinates for each OCEAN dimension and class: blue for low, green for medium, and red for high values.}
  \label{fig:xy}
\end{figure*}

To analyze the synthetic gaze trajectory data visually,   we plot the average trajectory for each personality dimension and class for 1000 generated data points spanning 300 time steps (5 seconds). For each time step in the 5-second window, we compute the average x and y values.  Figure~\ref{fig:xy} shows that gaze coordinates for each personality class are clearly distinguishable from each other.  These trajectories are consistent with some of the findings by Fukuyama et al. ~\cite{Fukuyama2002}.  They report that looking up is rated as being dominant, a trait of extroversion.  They find looking down to be associated with anxiety, which confirms the decreasing  values for our neuroticism data. Of course, perception of personality and its actual expression are different concepts, and not every cue is an indicator of personality. However, such clear associations lead the way for future research directions.

\begin{figure*}
  \includegraphics[width=\textwidth]{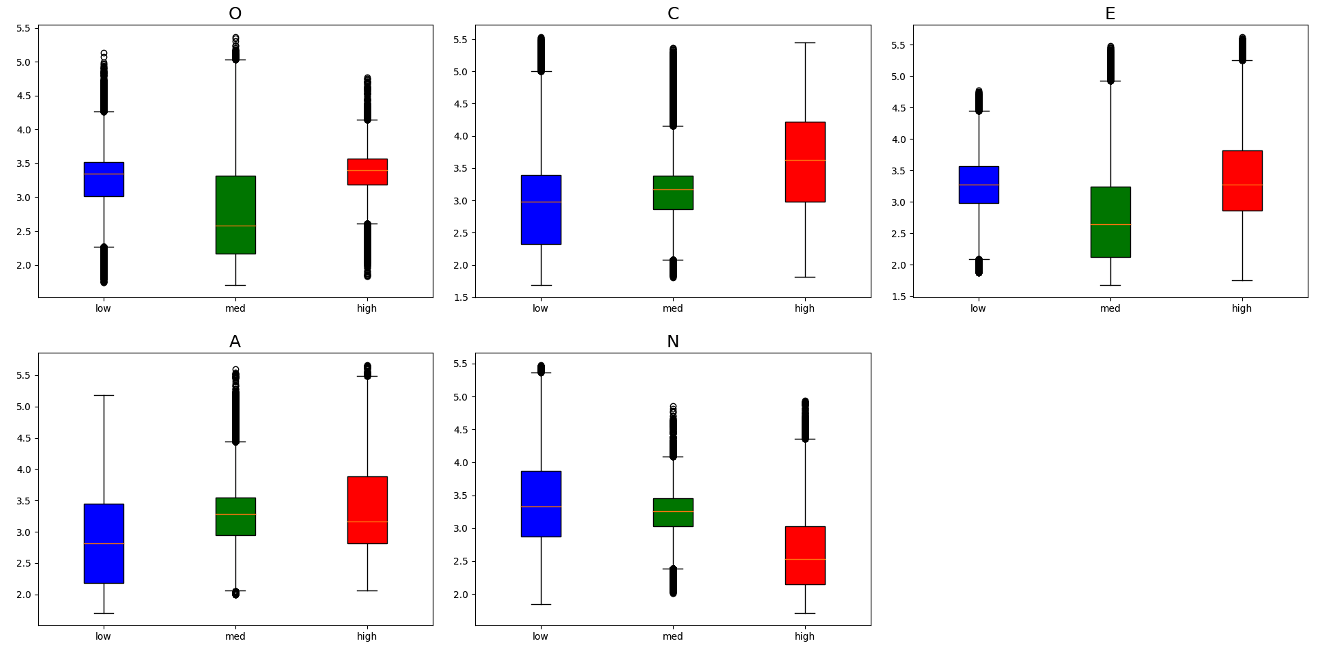}
  \caption{Synthesized pupil sizes for each OCEAN dimension and class: blue for low, green for medium, and red for high values.}
  \label{fig:pupils}
\end{figure*}

We also visualize the average pupil size for each personality dimension and class for each time step in Figure~\ref{fig:pupils}. Overall, larger pupil sizes seem to be associated with positive traits (considering emotional stability as the positive pole). 

Possibly due to the discrete nature of blinking features, the generated data for all the personality types converge to the same result.



\subsection{Gaze Animation}
\begin{figure}
\centering
    \includegraphics[width=0.9\textwidth]{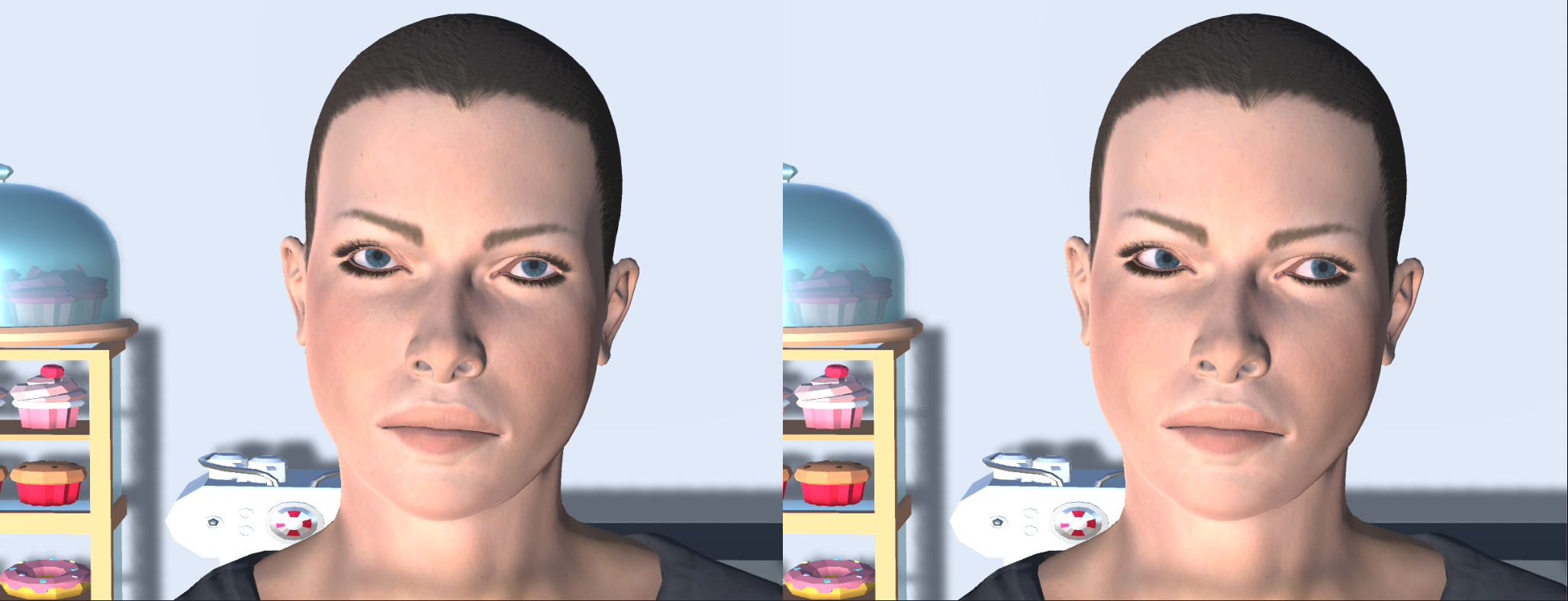}
  \caption{Animated gaze of an extrovert (left) vs. an introvert (right) model. }
  \label{fig:teaser}
\end{figure}

To animate eye movement, pupil dilation and blinking with respect to the data we created 3D human models by Adobe Fuse~\cite{AdobeFuse}. The model includes blend shapes for face, making it easy to animate facial expressions, and in our case, blinking motion. Figure~\ref{fig:teaser} shows the same humanoid model animated with gaze motions of different personalities.

The eye tracker glasses that were used to capture gaze data (SMI) have \ang{60} horizontal and \ang{46} vertical field of view angles. Using these angles, we convert the x and y values which are in the range $[0, 1]$ corresponding to the device space coordinates to the world space.
The target in the world space is the look-at direction of the eyes. For convenience, we take the middle point of the left and right eyes as the eye position. The target position is computed as follows:

\begin{eqnarray}
target_{world_x} &=& (2x - 1) d| tan(\ang{30})| + eye_x \nonumber \\ 
target_{world_y} &=& (2y - 1)  d| tan(\ang{23})| + eye_y \nonumber \\
target_{world_z} &=& d + eye_z \nonumber\\
\end{eqnarray}
, where $d$ is the maximum viewing distance. Note that the viewing direction from the eye position to the target point will be along the same line regardless of the viewing distance $d$, so this value will be canceled out later when computing the eyeball transforms. In addition to rotating the eyeballs to align with the look-at vector, we update the weights of the eyelid blendshapes so that they move naturally when the eyes move up and down.

\begin{figure}
  \includegraphics[width=0.5\textwidth]{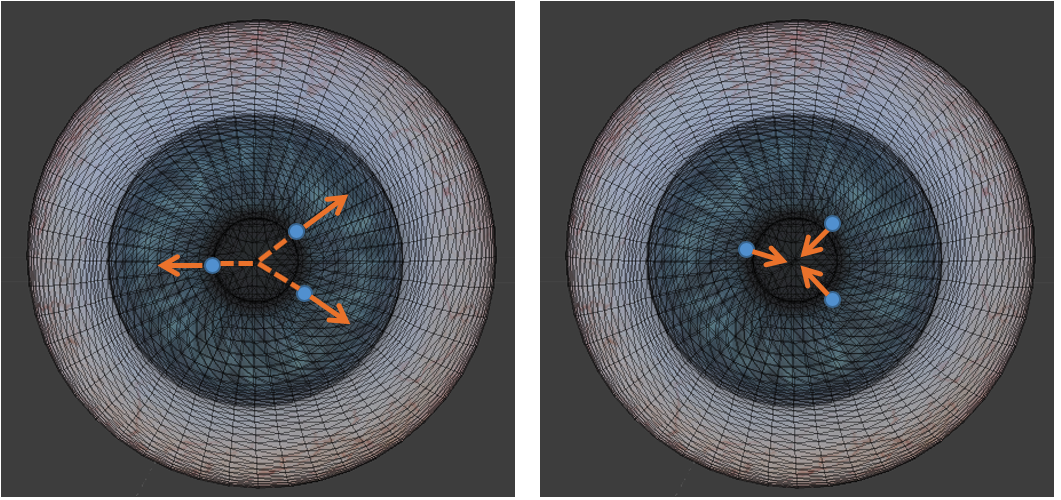}
  \caption{Animating pupil dilation.}
  \label{fig:eyeball}
\end{figure}

We  use a separate eyeball mesh to simulate the pupil dilation. We update the pupil dimension by applying forces to the vertices on the pupil perimeter towards or out of the center of the pupil as shown in Figure~\ref{fig:eyeball}.

\section{Conclusion}
We introduce a generative deep learning approach to synthesize time-series gaze data, and animate it on a virtual character. Our method is a preliminary step in this direction. The next step will be to create a personality-annotated gaze dataset during a conversation and to use our generative approach to this data. The data will include  similar features, but additionally, it will have information about the conversation target as well as head and torso pose information. We believe that the social nature of the task will help capture more salient features of personality expression.

\bibliographystyle{unsrtnat}
\bibliography{whole}

\end{document}